\documentclass[usletter, 10 pt, conference]{ieeeconf} 

\IEEEoverridecommandlockouts                              
\input{abkuerzung.sty}

\IEEEoverridecommandlockouts                              

\usepackage{amssymb,amsmath}
\usepackage{optidef}
\usepackage{textcomp}

\usepackage{algorithm}
\usepackage{lettrine}
\usepackage{siunitx}
\usepackage{url}
\usepackage{cite}
\usepackage{import}

\usepackage{multirow}

\usepackage{graphicx}
\usepackage{subcaption}
\usepackage{xcolor}
\usepackage{soul}
\usepackage[capitalise]{cleveref}

\global\long\def\dq#1{\underline{\boldsymbol{#1}}}%

\global\long\def\myvec#1{\boldsymbol{#1}}%

\newcommand{\anirban}[1]{{\textcolor{black}{#1}}}

\newcommand{\riddhi}[1]{{\textcolor{black}{#1}}}

\newcommand{\haowen}[1]{{\textcolor{black}{#1}}}

\usepackage{bm}
\makeatletter
\newcommand{\vast}{\bBigg@{4}}
\newcommand{\Vast}{\bBigg@{5}}
\makeatother

\title{\LARGE \bf On the Synthesis of Reactive Collision-Free Whole-Body Robot Motions: A Complementarity-based Approach}
\author{Haowen Yao$^{1\ast}$, Riddhiman Laha$^{1\ast}$, Anirban Sinha$^{2\ast}$, Jonas Hall$^{3}$, \\Luis F.C. Figueredo$^{4}$, Nilanjan Chakraborty$^{5}$, and Sami Haddadin$^{1}$
\thanks{$^\ast$Equal contribution. $^{1}$The authors are with the Munich Institute of Robotics and Machine Intelligence (MIRMI), Technical University of Munich (TUM), Munich, Germany. Email:\texttt{\{haowen.yao, riddhiman.laha, haddadin\}@tum.de}.$^{2}$Author is with GE global research, NY. Email: \texttt{anirban.sinha1@ge.com}.$^{3}$Author is with Boston University. Email: \texttt{hallj@bu.edu}.$^{4}$Author is with University of Nottingham. Email: \texttt{luis.figueredo@ieee.org}.$^{5}$Author is with Department of Mechanical Engineering, Stony Brook University. Email: \texttt{nilanjan.chakraborty@stonybrook.edu}.
This work is partially supported by the US Department of Defense through ALSRP under Award No. HT94252410098.
}
}
\renewcommand{\baselinestretch}{0.96}
\begin{document}

\maketitle
\begin{abstract}
\riddhi{This paper is about generating motion plans for high degree-of-freedom systems that account for collisions along the entire body. A particular class of mathematical programs with complementarity constraints become useful in this regard. Optimization-based planners can tackle confined-space trajectory planning while being cognizant of robot constraints. However, introducing obstacles in this setting transforms the formulation into a non-convex problem (oftentimes with ill-posed bilinear constraints), which is non-trivial in a real-time setting. To this end, we present the FLIQC (Fast LInear Quadratic Complementarity based) motion planner. Our planner employs a novel motion model that captures the entire rigid robot as well as the obstacle geometry and ensures non-penetration between the surfaces due to the imposed constraint. We perform thorough comparative studies with the state-of-the-art, which demonstrate improved performance. Extensive simulation and hardware experiments validate our claim of generating continuous and reactive motion plans at 1 kHz for modern collaborative robots with constant minimal parameters.}

\end{abstract}
\section{Introduction and State Of The Art}
%
Motion planning for robot manipulators in the presence of dynamic obstacles is a key capability that robots should have to ensure their safe operation in various applications ranging from manufacturing to home robotics~\cite{schwartz1988survey,wilfong1988motion,elbanhawi2014sampling}. A multitude of reactive motion planning approaches have been proposed in the literature that create velocity (or force) fields to avoid obstacles and also ensure progress towards the goal~\cite{khatib1986real,koditschek1987exact,rimon1992exact,laha2021reactive,laha2023predictive}. These velocity (vector) fields are primarily defined over the real Euclidean space; hence, they are inherently designed for point robots with point obstacles~\cite{becker2023motion}. Although these approaches have been extended to non-point rigid bodies, they require heuristic point placements to approximate a rigid body~\cite{becker2023informed}. These choices of the discrete set of points to represent the rigid bodies and the parameters for defining the fields can lead to artifacts related to local minima~\cite{sinha2023oc3}. Thus, in certain cases, the robot might not reach its goal and violate the collision constraints~\cite{dahlin2023creating}. Therefore, real-time reactive motion planning, that encodes robot and obstacle geometry, with a minimal parameter set remains an open challenge.

Keeping the same vision of trajectory generation, a practical non-real-time approach was undertaken in~\cite{ratliff2015understanding} where a robust local optimization was performed in order to find the path around obstacles. This theme was also explored earlier with a continuous global optimization module in CHOMP (gradient-based)~\cite{ratliff2009chomp} and STOMP (gradient-free)~\cite{kalakrishnan2011stomp}. The authors in TrajOpt~\cite{schulman2013finding}, a state-of-the-art collocation-based trajectory planner, formalize the problem as a non-linear program and utilize Sequential Quadratic Programming (SQP) to come up with valid paths. Nevertheless, the design leads to a quadratic increase in computation time as the number of waypoints and robot degree-of-freedom increases~\cite{sundaralingam2023curobo}.

Given the known physical fact that rigid objects are unable to interpenetrate, trying to create repulsive velocity or force fields is a computational approach to model interpenetration~\cite{slotine1988robust}. However, another approach to ensure non-penetration of rigid objects is the use of complementarity constraints, which is quite popular in the modeling and simulation of dynamics of rigid bodies in intermittent contact~\cite{posa2014direct,lidec2023contact,le2024fast}. Within the context of motion planning, complementarity constraints have been used to model collision avoidance in~\cite{chakraborty2009complementarity}, where the motion planner was designed to be a local planner within a global sampling-based motion planning framework~\cite{orthey2023sampling}. In this paper, we use a complementarity-based model to model collision avoidance with dynamic obstacles. This allows us to model whole-body collision avoidance without any approximation of the given geometry as well as without the need to tune parameters to set up artificial potentials or barriers to avoid collisions~\cite{singletary2019online,murtaza2021real,laha2021coordinated,murtaza2022safety,landi2019safety}. 
\begin{figure}[t]
    \begin{center}    \includegraphics[width=0.48\textwidth]{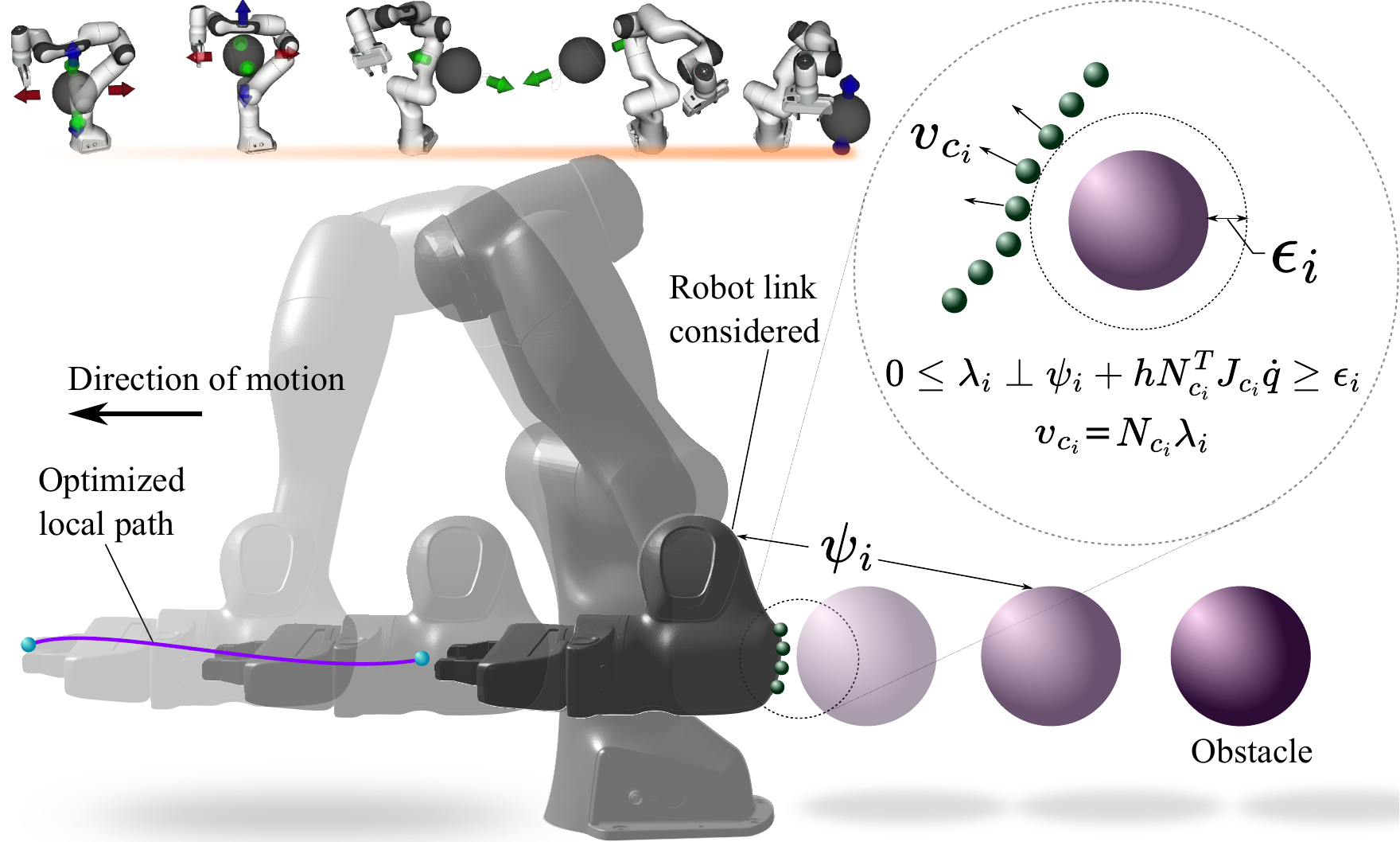}
    \caption{%
     Top left inset depicts the reactive behavior of our planner with an interactive obstacle. The main illustration highlights a scenario where the purple dynamic obstacle approaches the robot's wrist as it is moving towards a goal. As soon as the robot link (green spheres denote the tracked surface) enters the obstacle sphere of influence ($\epsilon_i$), it experiences a compensating velocity $\myvec{v}_{c_i}$ (magnified portion) that slides it along the surface,
     until it escapes. $\psi_i$ represents the robot link surface to the obstacle surface distance.
}
    \label{fig:cover_pic}
    \end{center}
    \vspace{-30pt}
\end{figure}
The complementarity constraint (Fig.~\ref{fig:cover_pic}) between two rigid objects is written as 
$ \lambda d = 0$, $\lambda \geq 0$, $d \geq 0$, where $d$ is the distance between the two objects, and $\lambda$ is usually interpreted as the force required to avoid interpenetration (or collision if we assume a safety distance threshold). It has been shown in~\cite{sinha2021task} that if the state update equation of the dynamical system is at the kinematic level, $\lambda$ can be interpreted as the velocity to be given to the robot (at the point of virtual contact) to avoid collisions as shown in Fig.~\ref{fig:cover_pic}. However, the velocity to avoid collision must be given at the joint level, and the offline formulation in~\cite{sinha2021task} does not model any optimization criterion or joint rate constraints. 

\noindent\textit{Contributions:} The key contribution of this paper is to formulate the local reactive collision avoidance problem as an optimization problem with whole-body collision avoidance modeled as a linear complementarity constraint. Although this is a non-convex optimization problem due to the presence of the complementarity constraints, we show that the problem can be solved in real-time by solving a sequence of convex optimization problems. We increase the flexibility of the traditional complementarity-based motion planner by simultaneously handling convex quadratic objective terms. We evaluate the capability of our approach to respect joint limits and obstacles in various scenarios with a $7$ DoF collaborative robot and compare with the state-of-the-art. Lastly, we open-source our ROS-based planner for the community\footnote{The 
 repository can be accessed at 
\url{https://github.com/hwyao/FLIQC-planner}.}.

\section{Background and Problem Formulation}\label{sec:prob_def}
In this section, we present the requisite background and formulation of the motion planning problem with collision avoidance modeled as a Linear Complementarity Problem (LCP) that encapsulates the robot and obstacle geometry.
\subsection{Distance Function}
Consider robot links and obstacles 
modeled as 
rigid bodies, 
each denoted by $\Omega$, 
with a closed and bounded subset of ${\mathbb R}^3$. 
Let $\myvec{x}\in\mathrm{SE(3)}$ be the pose 
of a rigid body in some world frame~\cite{murray2017mathematical}. The set of points occupied by the object changes with its pose $\myvec{x}$ and therefore, in general, the set of points of the rigid body is a function of $\myvec{x}$, which we will denote as $\Omega(\myvec{x})$. 
During implementation, the geometry of these objects (e.g., robot links and obstacles) can be represented using primitives, polytopes, or point clouds~\cite{kim2014semantic,behl2019pointflownet,padilla2020survey,eckhoff2023towards}.

For any pair of rigid bodies $(\Omega_j, \Omega_k)$, we define a distance function $\psi(\Omega_j,\Omega_k) \in \mathbb{R}^{+}$, which is $0$ if $\Omega_j \cap \Omega_k \neq \emptyset$ (i.e., if the objects touch or overlap) and positive otherwise. Note that the Euclidean distance satisfies the properties of the distance function defined above. The algebraic distance can also be used as a distance function, as shown in~\cite{chakraborty2009optimization}.


\subsection{Kinematic and Geometric Motion Model}\label{sec:model_def}
We consider a $n$-degree-of-freedom (DoF) serial chain manipulator. Let $\myvec{q}\in\mathbb{R}^{n}$ be the configuration of the robot with the $j$th component of $\myvec{q}$, denoted by $q_j$, the $j$th joint displacement. The notation $\myvec{q}_{1:j}$ denotes the subvector of $\myvec{q}$ consisting of the first $j$ joint displacements. Let $\myvec{x}_j \in \mathrm{SE(3)}$, $ 1 \leq j \leq n$ be the pose of the $j$th link of the robot associated with the $j$th joint. Using the forward position kinematics map, the pose of the $j$th link is given by $\myvec{f}_j: \mathbb{R}^{j} \rightarrow \mathrm{SE(3)}$, 
\begin{equation}\label{eq:dir_kin1}
    \myvec{x}_j = \myvec{f}_j(\myvec{q}_{1:j}), \qquad j = 1, \cdots, n.
\end{equation}
Therefore, the position of the points comprising the link $j$ depends on the joint configuration up to the link $j$, $\myvec{q}_{1:j}$, and can be denoted as $\Omega_j(\myvec{f}_j(\myvec{q}_{1:j}))$ -- for brevity we will use $\Omega_j(\myvec{f}(\myvec{q}))$. We will denote obstacles by $\Omega_k(\myvec{x}_k)=\Omega_k(\myvec{x})$ where $\myvec{x}_k \in \mathrm{SE(3)}$ is the pose of obstacle $k$.


Let $\myvec{\zeta} = [\myvec{v}^T, \bm{\omega}^T]^T$ be the twist (generalized velocity) of the end effector of the manipulator, with $\myvec{v}\in\mathbb{R}^3$ denoting the linear velocity, and $\bm{\omega}\in\mathbb{R}^3$ being the angular velocity of the manipulator end-effector, respectively. The differential kinematics relationship between the joint rates and end-effector twist are given by~\cite{lynch2017modern},
\begin{equation}\label{eq:dir_kin2}
    \myvec{\zeta} = \myvec{J}(\myvec{q})\dot{\myvec{q}}.
\end{equation}
where $\myvec{J}(\myvec{q})$ is the manipulator Jacobian. In the subsequent discussion, we will also use a unit dual quaternion representation of $\mathrm{SE(3)}$, which we have used in our implementation~\cite{adorno2017robot}. We will use $\dq{x}$ to represent the pose of a body with unit dual quaternions. The differential kinematics relationship then is $\dot{\dq{x}} = \myvec{J}(\myvec{q})\dot{\myvec{q}}$, where, now the Jacobian $\myvec{J}(\myvec{q})$ also contains the representation/analytical Jacobian~\cite{figueredo2013robust}.



\subsection{Collision Avoidance Constraint}
Using the distance function and the notation introduced above, for link $j$ and obstacle $k$, the collision avoidance constraint can be written as,
\begin{equation}\label{eq:non-pen-ineq}
    \psi \left( \Omega _j\left( \myvec{f}\left( \myvec{q} \right) \right) ,\Omega _k\left( \myvec{x} \right) \right) \triangleq \psi _i\ge \epsilon_i, 
\end{equation}
where $\epsilon_i$ is a safety threshold for a particular instance $i$.



\haowen{During time $t\in[0,T]$ obstacle information can be obtained using some desired sensor resulting in trajectories $\Omega_k(\myvec{x}(t))$. Formally, we seek a collision-free joint trajectory $\myvec{q}(t)$  that commands the robot to execute some main task (e.g., with a specific start and goal) while each rigid link-obstacle pair satisfies the non-penetration condition along the entire surface. Here, $\psi_i$ represents the closest distance between the link and obstacle at the $i$th instance.}

\subsection{Collision Avoidance via Complementarity Constraints}
From the physical perspective, as the point of interest in the robot structure enters the sphere of influence of the obstacle, the rigid robot link should experience a contact compensating velocity $\myvec{v}_{c}\in\mathbb{R}^3$ along the normal direction of virtual contact to prevent penetration (see Fig.~\ref{fig:cover_pic}). At the same instant, we want this dynamic equilibrium of impenetrability to hold as the robot link slides along the sphere's surface until it escapes the sphere of influence, and this \anirban{(virtual) contact compensating velocity} reduces to zero. This desired behavior can be modeled as a switched system, in which the contact compensating velocity is forced to be zero whenever the distance function exceeds a certain threshold. We can encode this switch mathematically by introducing the complementarity constraint 
\begin{equation}\label{eq:comp_part}
    0 \leq \lambda_i \perp \psi_i - \epsilon_i \geq 0,
\end{equation}
where $i$ denotes the instance of the considered link and obstacle pair and $\lambda_i \in \mathbb{R}^+$ is a factor that scales the motion's unit normal vector $\myvec{N}_{c_i} \in \mathbb{R}^3$. As a result, we obtain $\myvec{v}_{c_i} = \myvec{N}_{c_i}\lambda_i$. 
The notion of complementarity in~\cref{eq:comp_part} compactly captures non-negativity and orthogonality:
\begin{equation*}
    \begin{split}
        0 \leq \lambda_i,\quad \lambda_i (\psi_i - \epsilon_i) = 0,\quad \psi_i - \epsilon_i \geq 0,
    \end{split}
\end{equation*}
and is thus non-linear and non-convex due to the bi-linear equality constraint. Combining the robot kinematics (\ref{eq:dir_kin1}),(\ref{eq:dir_kin2}) with the obstacle avoidance constraints (\ref{eq:comp_part}), we can formulate a whole-body collision-free planning method as a Differential Complementarity Problem (DCP)
\begin{equation}\label{eq:DCP_comp_1}
    \begin{cases}
   \dot{\myvec{q}}=\myvec{J}^{\dag}\dot{\dq{x}}+\sum {\myvec{J}_{c_i}^{\dag}\myvec{N}_{c_i}\lambda_i}\\
        0 \leq \myvec{\lambda} \perp \myvec{\psi} - \myvec{\epsilon} \geq 0
    \end{cases}\,,
\end{equation} 
where $\myvec{J}^{\dag}$ is the pseudo-inverse of the Jacobian, $\myvec{J}_{c_i}$ is the \anirban{Jacobian up to the point on the $i^{th}$ link that is closest to any obstacle}, and $\myvec{N}_{c_i}$ is the \anirban{unit} normal vector along contact point pair between the link and the obstacle. $\myvec{J}_{c_i}$ and $\myvec{N}_{c_i}$ are computed for each link and its corresponding closest obstacle pair. In the above equations, the vectorized complementarity constraints are to be read as component-wise complementarity constraints. DCPs are a subclass of problems in the general Mathematical Programs with Complementarity Constraints (MPCC) family, and are usually solved as a feasibility problem using an Euler time-stepping scheme~\cite{stewart1996implicit}, and the resultant system of equations is linearized using standard first-order approximations~\cite{sinha2021task}. This linearized form is conventionally known as Linear Complementarity Problem (LCP)~\cite{cottle2009linear}. \haowen{For LCP systems,} given a vector $\myvec{u}\in\mathbb{R}^{n}$ and a matrix $\myvec{G}\in\mathbb{R}^{n\times n}$, the LCP$(\myvec{u},\myvec{G})$ describes the program:
\begin{equation}\label{eq:lcp}
\begin{aligned}
& {\text{find}}
& & \myvec{\lambda} \in \mathbb{R}^{n} \\
& \text{subject to}
& & 0 \leq \myvec{\lambda} \perp \myvec{G}\myvec{\lambda} + \myvec{u} \geq 0.
\end{aligned}
\end{equation}
This LCP may have multiple or no solutions.

\riddhi{Note that the DCP in \cref{eq:DCP_comp_1} or the more general LCP conceptualization in \cref{eq:lcp} does not include any objective during the feasibility checks at each instant. In practical scenarios, robots would need to find the shortest, minimum energy, or safest paths while executing various other tasks like obstacle avoidance, singularity escape, or joint-limits circumvention. We elucidate how to tackle this problem by incorporating a general objective in the next subsection.}

\subsection{Problem Formulation}
\riddhi{We are now equipped to formalize the problem. The main objective is to develop a fast, efficient, and local task-space planner that ensures whole-body collision avoidance. %
} 

\noindent \textit{Formally, given task-space twist coordinates $\dot{\dq{x}}\in\mathbb{R}^8$ that we need to track, and a convex quadratic cost function $\bm{F}$, we seek the instantaneous joint velocities $\dot{\myvec{q}}\in\mathbb{R}^n$ (within the permissible mechanical bounds of the actuators), and the multipler $\lambda_i$, such that the obstacle avoidance constraints are respected. If no such joint velocity is feasible, \anirban{ return $\emptyset$}.}

\noindent \textbf{Mathematical Problem Statement} 
\begin{mini!}
{\dot{\myvec{q}}\in\mathbb{R}^n}{\frac{1}{2}{\myvec{\dot{q}}}^{T}\myvec{Q}{\myvec{\dot{q}} + {\myvec{c}}^{T}\myvec{\dot{q}} }}
{\label{eq:lcqp_full}}{}
\addConstraint{\myvec{\dot{q}} = \myvec{J}^{\dag}\dot{\dq{x}} + \sum\myvec{J}_{c_i}^{\dag}\myvec{N}_{c_i}\lambda_i}\label{eq:lcqp_kin}
\addConstraint{0 \leq \lambda_i \perp \psi_i + h\myvec{N}_{c_i}^{T}\myvec{J}_{c_i} \myvec{\dot{q}}\geq \epsilon_i}\label{eq:lcqp_comp_cond}
\addConstraint{\myvec{\dot{q}}^{-}\leq \myvec{\dot{q}}\leq \myvec{\dot{q}}^{+}}.\label{eq:joint_vel_limit}
\end{mini!}
where $\myvec{Q}\in\mathbb{R}^{n\times n}$ is a symmetric \anirban{positive definite} matrix, which can allow us to model cost criteria like kinetic energy. The quantity $\psi_i + h\myvec{N}_{c_i}^{T}\myvec{J}_{c_i} \myvec{\dot{q}}$ is a first-order approximation of the distance at the next time step, and $h$ is the step-size. The objective of Eq~\eqref{eq:lcqp_full} is the cost function $\bm{F}:\mathbb{R}^{n}\rightarrow\mathbb{R}$.

\section{Fast Linear Quad. Comp. (FLIQC) Program} \label{sec:FLIC}

In this part, we illustrate how our proposed algorithm effectively handles collisions along with an additional quadratic objective and generates optimal joint velocities.

\subsection{Solution Approach}
The problem considered in (\ref{eq:lcqp_full}) falls in the class of LCQPs, a notoriously difficult class of non-linear and non-convex optimization problems with non-smooth feasible sets. To briefly discuss our approach, consider the generic LCQP  
\begin{mini!}
	{\myvec{y}\in \mathbb{R}^n}{\frac{1}{2} \myvec{y}^T \Tilde{\myvec{\bm{Q}}} \myvec{y} + \myvec{g}^T \myvec{y} \label{min:LCQP:intro:obj}}
	{\label{min:LCQP:intro}}{}
	\addConstraint{\myvec{b}}{\leq \myvec{A} \myvec{y} \label{min:LCQP:intro:A}}
	\addConstraint{\myvec{0}}{\leq \myvec{L}\myvec{y} \perp \myvec{R}\myvec{y} \geq \myvec{0}, \label{min:LCQP:intro:complementarity}}
\end{mini!}
where $\Tilde{\myvec{\bm{Q}}} \succ \bm{0}$ is positive definite, $\myvec{A}$ and $\myvec{b}$ describe linear constraints (including bounds on $\bm{y}$), and $\myvec{L}$ and $\myvec{R}$ describe the complementarity constraints. Note that every LCQP can be represented in this form. The vectorized complementarity constraints in \cref{min:LCQP:intro:complementarity} are to be read pair-wise, meaning that each row of $\myvec{L}\myvec{y}$ must be complementary to the respective row of $\myvec{R}\myvec{y}$. The constraint \eqref{min:LCQP:intro:complementarity} is equivalent to the set of constraints $\myvec{0} \leq \myvec{L}\myvec{y}, \myvec{R}\myvec{y}$ and $\myvec{y}^T\myvec{L}^T \myvec{R}^T \myvec{y} = 0$, which are referred to as non-negativity and orthogonality constraints, respectively. Note that the orthogonality constraint is now one-dimensional, so the original problem without this constraint is a convex QP. Thus, a common and well-understood approach to solving \eqref{min:LCQP:intro} consists of penalizing the orthogonality constraint. This results in the non-convex QP 
\begin{mini!}
	{\myvec{y} \in \mathbb{R}^n}{\frac{1}{2} \myvec{y}^T \Tilde{\myvec{\bm{Q}}} \myvec{y} + \myvec{g}^T \myvec{y} + \rho\myvec{y}^T\myvec{L}^T \myvec{R}^T \myvec{y}  \label{min:LCQP:penalty:obj}}
	{\label{min:LCQP:penalty}}{}
	\addConstraint{\myvec{b}}{\leq \myvec{A} \myvec{y} \label{min:LCQP:penalty:A}}
	\addConstraint{\myvec{0}}{\leq \myvec{L}\myvec{y}, \myvec{R}\myvec{y}, \label{min:LCQP:penalty:complementarity}}
\end{mini!}
where $\rho >0$ is a penalty parameter. It was shown that this penalty reformulation is exact for a sufficiently large but finite value of $\rho$~\cite{ralph2004some}. To avoid ill-conditioning of the problem, a sufficient penalty parameter is typically found via a homotopy approach~\cite{nurkanovic2020limits}. The recently introduced solver LCQPow~\cite{hall2022lcqpow}, implements the penalty-based sequential convex programming method originally introduced in~\cite{hall2021sequential}. LCQPow addresses the LCQP \eqref{min:LCQP:intro} via this reformulation. To solve the non-convex penalty reformulation, it performs a sequence of convex subproblems that are obtained from linearizing the penalty function. Therefore, we are able to solve \eqref{eq:lcqp_full} in real-time by employing this solver. 

In the context of our problem \eqref{eq:lcqp_full}, the decision variable of \eqref{min:LCQP:penalty:obj}, $\bm{y}$ is the concatenated vector of $\dot{\bm{q}}$ and $\lambda_i$'s. The matrix $\bm{A}$ and vector $\bm{b}$ can be formed from \eqref{eq:lcqp_kin} and \eqref{eq:joint_vel_limit}. Finally, the $\bm{L}$ and $\bm{R}$ sparse matrices in \eqref{min:LCQP:penalty:complementarity} can be constructed from \eqref{eq:lcqp_comp_cond}. If we define the number of links as $n_l$ and the number of potential contacts as $n_c$, then, 
    $\bm{y} = \left[\dot{\bm{q}}, \lambda_{1}, \dots, \lambda_{n_c}\right]^T \in \mathbb{R}^{n_l+n_c}$,
    $\bm{Ly} = \left[\lambda_{1}, \dots, \lambda_{n_c}\right]^T \in \mathbb{R}^{n_c}$,
    $\bm{Ry} = \left[\psi_1 + h\myvec{N}_{c_1}^{T}\myvec{J}_{c_1}\dot{\bm{q}}, \dots , \psi_{n_c} + h\myvec{N}_{c_{n_c}}^{T}\myvec{J}_{c_{n_c}}\dot{\bm{q}}\right]^T \in \mathbb{R}^{n_c},$
    $\bm{Ay} = \left[\dot{\bm{q}} - \sum {\myvec{J}_{c_i}^{\dag}\myvec{N}_{c_i}\lambda_i} ;\dot{\bm{q}};-\dot{\bm{q}}\right] \in \mathbb{R}^{3n_l}$,
    and $\bm{b} = \left[\myvec{J}^{\dag}\dot{\dq{x}}; \myvec{\dot{q}}^{-} ; -\myvec{\dot{q}}^{+}\right]$ (the notation $;$ denotes row arrangement).

Looking at problem \eqref{eq:lcqp_full}, we find that safety is guaranteed, even if FLIQC terminates prematurely, i.e., if it terminates with an iterate that violates the orthogonality condition \eqref{eq:lcqp_comp_cond}. Any feasible point from \eqref{min:LCQP:penalty} satisfies $\psi_i + h\myvec{N}_{c_i}^{T}\myvec{J}_{c_i} \myvec{\dot{q}}\geq \epsilon_i$, which ensures safety. We now show that any iterate of FLIQC lies in the feasible set of \cref{min:LCQP:penalty}. The initial iterate is computed by solving \eqref{min:LCQP:penalty} with $\rho=0$, which is the convex QP that arises when removing the orthogonality from~\eqref{eq:lcqp_comp_cond}. Any future iterate of FLIQC is obtained through solving problems with a modified objective function (by increasing the penalty parameter and linearizing the penalty function), but the feasible set always coincides with that of~\eqref{min:LCQP:penalty}. Premature termination, e.g., triggered through a time out, thus leads to safe but potentially suboptimal/conservative solutions. This feature also enables it to work as a anytime-planner. Compared to PATH, which can only find a feasible point of the DCP~\eqref{eq:DCP_comp_1}, FLIQC has the capability of additionally optimizing a convex quadratic performance metric. Additionally, PATH exploits (stabilized) Newton's method for the interior point problem and takes significant time to converge~\cite{drumwright2012extensive}.

\subsection{Implementation Details}
We make use of the $7$~DoF Franka Robot as our test-bed. During simulation in \cref{sec:simulation}, the position and velocity kinematics are 
computed using the DQRobotics library~\cite{adorno2020dq}. In contrast, the hardware experiments of \cref{sec:exp_results} use libraries provided by the manufacturer for stable performance.

To facilitate rapid distance computation between robot links and obstacles, we employed basic geometric primitive shapes for obstacle modeling. Obstacles are modeled as spheres, and the robot itself is modeled as a series of cylinders connecting the link frames\footnote{Although spheres are our choice herein, FLIQC accepts any convex primitive. The only change required would be the distance computations.}. The safety padding distance around the robot is defined as $15$ cm. This implementation enables us to efficiently expedite contact point calculations by conducting line-sphere calculations. For straightforward extensions to more complex shapes, we refer readers to~\cite{montaut2024gjk++}. Another point to be highlighted is that the obstacle coordinate only contains position. Correspondingly in \cref{eq:lcqp_full}, $\myvec{J}_{c_i}$ is the position Jacobian up to the contact point, and $\myvec{N}_{c_i}$ is the normal vector along the fastest separation direction. 

\noindent\textit{Cost function}: We currently apply minimum energy control cost~\cite{halevi2014minimum}. In other words, we reformulate the robot's rigid body kinetic energy as a squared velocity norm $\mathcal{K}(\myvec{q},\dot{\myvec{q}}) = \frac{1}{2}\dot{\myvec{q}}^T\myvec{M}(\myvec{q})\dot{\myvec{q}}$, which we intend to minimize at each step. Here, $\myvec{M}$ is the inertia matrix of the manipulator. Although an important quantity in motion, kinetic energy is not usually modeled in kinematic motion optimization strategies like CHOMP, STOMP, and TrajOpt. Further extensions are possible by employing different resolution schemes and choice of $\mathcal{K}$. Analysis of such schemes is deferred to future work. With regard to the optimization cost, the current implementation uses the identity matrix $\myvec{M}(\myvec{q})=\myvec{I}$, thereby directly optimizing for joint velocities. Warm starting for the optimization solver is also performed for fast convergence at each step.


\section{Illustrative Planar Example} \label{sec:illus_exam}
\begin{figure}[t]
    \begin{center}    \includegraphics[width=0.39\textwidth]{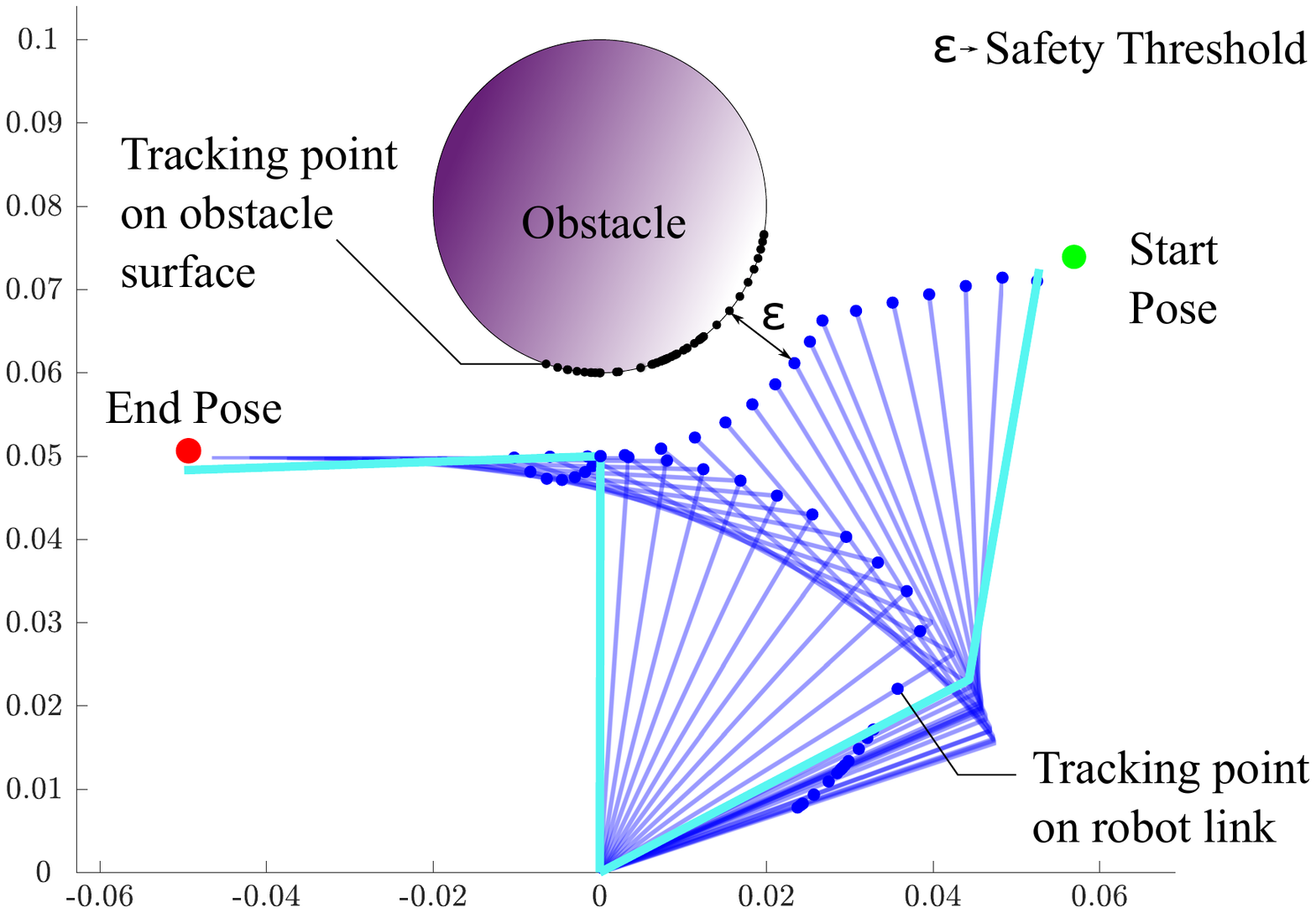}
    \caption{%
    Evolution of obstacle avoidance  using FLIQC for a planar $2$R robot. The cyan color represents the starting and goal configurations of the robot. 
}
    \label{fig:illust_example}
    \end{center}
    \vspace{-25pt}
\end{figure}

To better illustrate the proposed strategy and the overall idea, let us consider a planar $2$R ($2$ revolute joints) manipulator with equal link lengths of \SI{0.05}{\metre} performing a reaching task in the $xy$ plane as shown in Fig.~\ref{fig:illust_example}. The starting configuration $\myvec{q}$ for the system is [$0.523,0.785$]rad, and the initial EE pose is at [$0.0563,0.0733$]m. The goal $\myvec{p}_g$ is set at [$-0.05,0.05$]m. An obstacle with a radius of \SI{0.02}{\metre} and a safety threshold of $\epsilon=$ \SI{0.01}{\metre} is placed at [$0.0,0.08$]m. The threshold is incorporated so as to define a \textit{detection shell} around the obstacle. This ensures that the avoidance mechanism kicks in only when a part of the robot structure enters this shell. 
We define the Cartesian velocity vector $\dot{\dq{x}}_p$, a proportional signal that drives the system to the goal as
\begin{equation}
    \dot{\dq{x}}_p = \frac{k_d\left( \myvec{p}_g - \myvec{p}_c \right)}{\|\myvec{p}_g - \myvec{p}_c \|} .
\end{equation} 
where $\myvec{p}_c$ is the current end-effector position and $k_d$ a positive scalar gain.
The joint angle vector $\myvec{q}$ and its rate of change $\dot{\myvec{q}}$ form the state space of the system. At each step of the planning process, we compute the distance $\psi$ between each link and the obstacle surface. Then, we solve the quadratic program involving \eqref{eq:lcqp_kin}- \eqref{eq:lcqp_comp_cond} with an iteration budget of $1000/$step. The step parameter $h$ is set to $0.001$.
Thus, we ensure through this program that for the \textit{i}th robot body, at each potential (virtual) contact, the product of the magnitude of the normal contact velocity $\myvec{v}_{c}$ and the distance $\psi$ is always zero. This can be matched in Fig.~\ref{fig:illust_example}, where the tip of the robot slides along the obstacle surface while respecting the safety barrier $\epsilon$. The generated joint velocities $\dot{\myvec{q}}$ from the QP can then be integrated to get positions ${\myvec{q}}$ for the next step as ${\myvec{q}_{t+1}} = \myvec{q}_t + h\dot{\myvec{q}}$. Termination of the entire program occurs when the goal threshold has been attained.     
\section{Comparative Simulation Analyses} \label{sec:simulation}
\begin{table}[t]
\centering
    \caption{Comparison with SOTA planners in various \textbf{static} scenarios. Note that FLIQC is the only real-time planner here.}
    \label{table:planner_sota_comp}
\begin{tabular}{c|c|c|c|c} 
  \bfseries Scenes & \bfseries Planner & \bfseries Success & \bfseries Path Len  & \bfseries Avg. Joint Mov. \\ 
    &   &   & [m] & [1e-3 rad]\\ 
 \hline
 Static    & CHOMP & 73/100 & 1.1604 & 2.4242 \\
    env.   & Bi-RRT & 79/100 & 0.85292 & 1.7611 \\
    1    & FLIQC & 67/100  & 0.63341 & 1.7121\\
  \hline
   Static  & CHOMP &  72/100 & 0.97265 & 2.2152 \\
     env.   & Bi-RRT & 66/100 & 0.78342 & 1.9457 \\
    2 & FLIQC & 78/100  & 0.59686  & 1.6501 \\
  \hline
   Static  & CHOMP & 75/100 & 1.0228 & 2.2862 \\
     env.   & Bi-RRT & 64/100 & 0.70887 & 1.6314 \\
    3 & FLIQC & 69/100  & 0.68569 & 1.7014 \\
  \hline
   Static  & CHOMP & 82/100 & 1.1928 & 2.4984 \\
     env.  & Bi-RRT & 80/100 & 0.95748 & 2.2932 \\
    4 & FLIQC & 77/100  & 0.60845 & 1.6921\\
  \hline
     Static  & CHOMP & 88/100 & 1.2489 & 2.6450 \\
     env.  & Bi-RRT & 77/100 & 0.83163 & 2.1577 \\
     5 & FLIQC & 87/100  & 0.61895 & 1.7305\\
  \hline
            & CHOMP & 390/500 & 1.1195 & 2.4138  \\
     Total  & Bi-RRT & 366/500 & 0.8269 & 1.9578 \\
            & \bf{FLIQC} & \bf{378}/500  &  \bf{0.6287} & \bf{1.6972}\\
  \hline
  \end{tabular}
    \vspace{-25pt}
\end{table}
In this section, we analyze in detail the performance of our proposed FLIQC planner with (a) the trajectory optimization-based planner CHOMP~\cite{ratliff2009chomp} and (b) the widely-used offline configuration space planner Bi-RRT~\cite{lavalle2001randomized}.
We quantitatively compare the performances of FLIQC to CHOMP and Bi-RRT in $5$ different scenarios with $100$ random feasible initial start and goal. $3$ static obstacles with different arrangements are deployed in the $5$ scenes. To prevent the experiments from consuming too much time, we set an interrupting condition for these planners. The FLIQC plannner can run a maximum $1500$ iterations before it is considered as failure. Meanwhile, the CHOMP and Bi-RRT planners with MATLAB default parameters have a maximum of $20$ seconds for optimization. The metrics of success rate, path length, and average joint movement per joint are considered as effective criteria to judge the performance. As observed in \cref{table:planner_sota_comp}, while the specific performances varies from scene to scene, the FLIQC planner generally has better path length $(0.6287~\textrm{m})$ and average joint movement $(0.00169~\textrm{rad})$. Although CHOMP has a marginally better success rate than FLIQC over the $500$ runs, it is not necessarily better when it comes to reactive behaviors. We also note that one possible question that arises is that Bi-RRT being a global planner has the lowest succcess rate. This is possibly due to the fixed (limited) iteration budget that we follow for the comparison.

\section{Hardware Experiments} 
\label{sec:exp_results}
This portion seeks to explicate the observations and generated data from a hardware-in-the-loop setting. The computer used for sending control commands in a $1$ kHz loop had a configuration of Intel(R) Core(TM) i$5$-$7600$ CPU @ $3.50$GHz having $4$ cores. The low-level controller, programmed in C++ and not fully optimized for efficiency, commands the velocity interface of the robot. 

It is important to note here that the parameters for FLIQC \textit{remain constant throughout all the scenarios}, as shown in \cref{table:param}. Specific parameter tuning would slightly influence some performances (e.g. solver running time, end-effector speed), but the safety distance is always maintained without fine-tuning these values. Regardless, we report the entire parameter set for reproduction purposes.

The following parameters are connected with eq. (\ref{eq:lcqp_full}) and influence the behavior of FLIQC: (i) $\|\myvec{\dot{x}}\|$ is the main task velocity, (ii) $\myvec{v}_{c_i}$ is the complementarity velocity, (iii) $\myvec{\dot{q}}$ is multiplied by the joint velocity factor $k_q$ before being sent as the input velocity control command. To achieve stability around singularities, the robust pseudo-inverse of Jacobian is applied and, therefore, requires a parameter for the damping of Jacobian~\cite{chiaverini1994review}. Furthermore, there are some additional parameters to improve performance. The parameter $\ell$ controls which link(s) we consider for the collision checker. In our experiments only links $3$, $4$, and $7$ are being investigated for collisions. Meanwhile, to improve upon real-time performance, we neglect obstacles beyond a certain threshold. In particular, when an obstacle is \SI{0.02}{\meter} (with the additional padding of \SI{15}{\cm}) away from the link, it is safely disregarded.

We consider $3$ distinct scenarios for a complete hardware evaluation of our proposed framework. This demonstrates the real-time evasive capabilities of FLIQC in highly constrained static scenes as well as in fast-changing environments. Further, we note that the raw output from the optimizer without any post-processing step is supplied to the real system.
\vspace{-10pt}
\begin{table}[ht]
\centering
    \caption{Parameter set for real robot experiments}
    \label{table:param}
\begin{tabular}{c|ccccc} 
  \bfseries Scene & \bfseries Main task & \bfseries Complementarity & \bfseries Joint & \bfseries Jacobian \\ 
   & \bfseries velocity & \bfseries velocity factor &  \bfseries velocity factor & \bfseries damping\\ 
 \hline
 Scn. $1$ & 0.2  & 0.05 & 0.2  & 0.001 \\
  \hline
    \end{tabular}
    \vspace{-15pt}
\end{table}
\subsection{Scenario \#1: Constrained Static Environment}

\begin{figure}[t]
    \begin{center}    \includegraphics[width=0.29\textwidth]{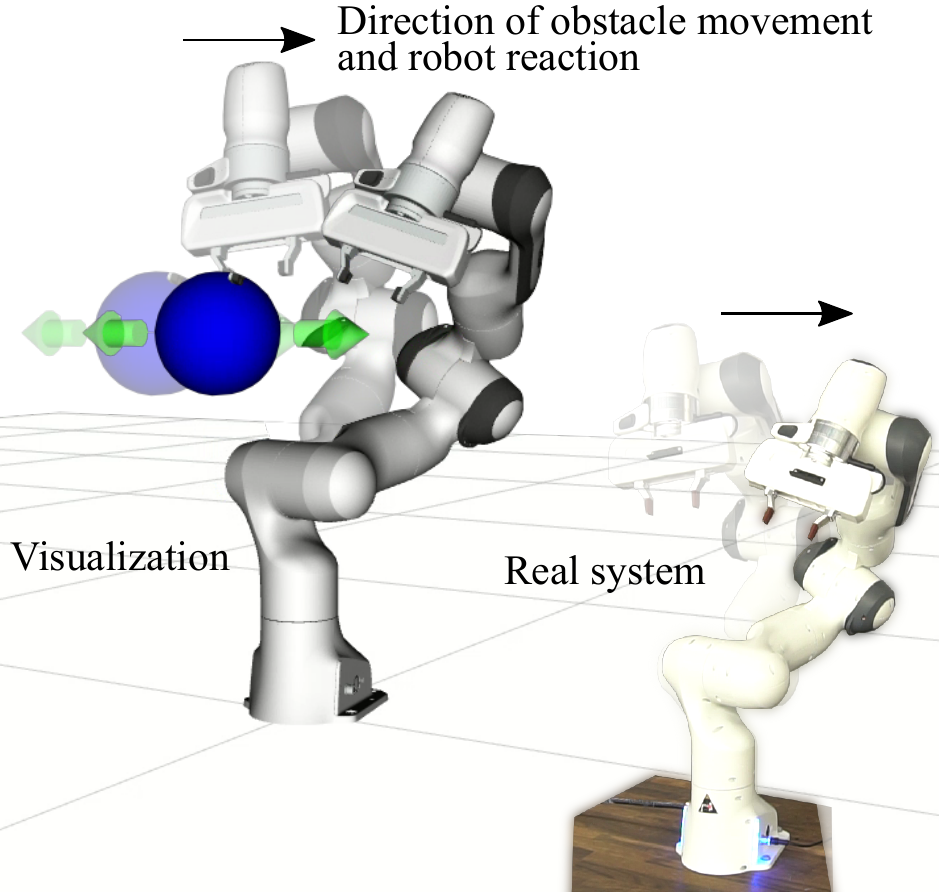}
    \vspace{-5pt}
    \caption{%
    Interactive marker scenario with hardware-in-the-loop: The robot reacts in real-time to the marker nudge from left to right at the elbow. 
}
    \label{fig:inter_marker}
    \end{center}
    \vspace{-25pt}
\end{figure}

\begin{figure}[ht]
    \centering
    \begin{subfigure}[b]{0.45\textwidth}
        \centering
        \includegraphics[width=\textwidth]{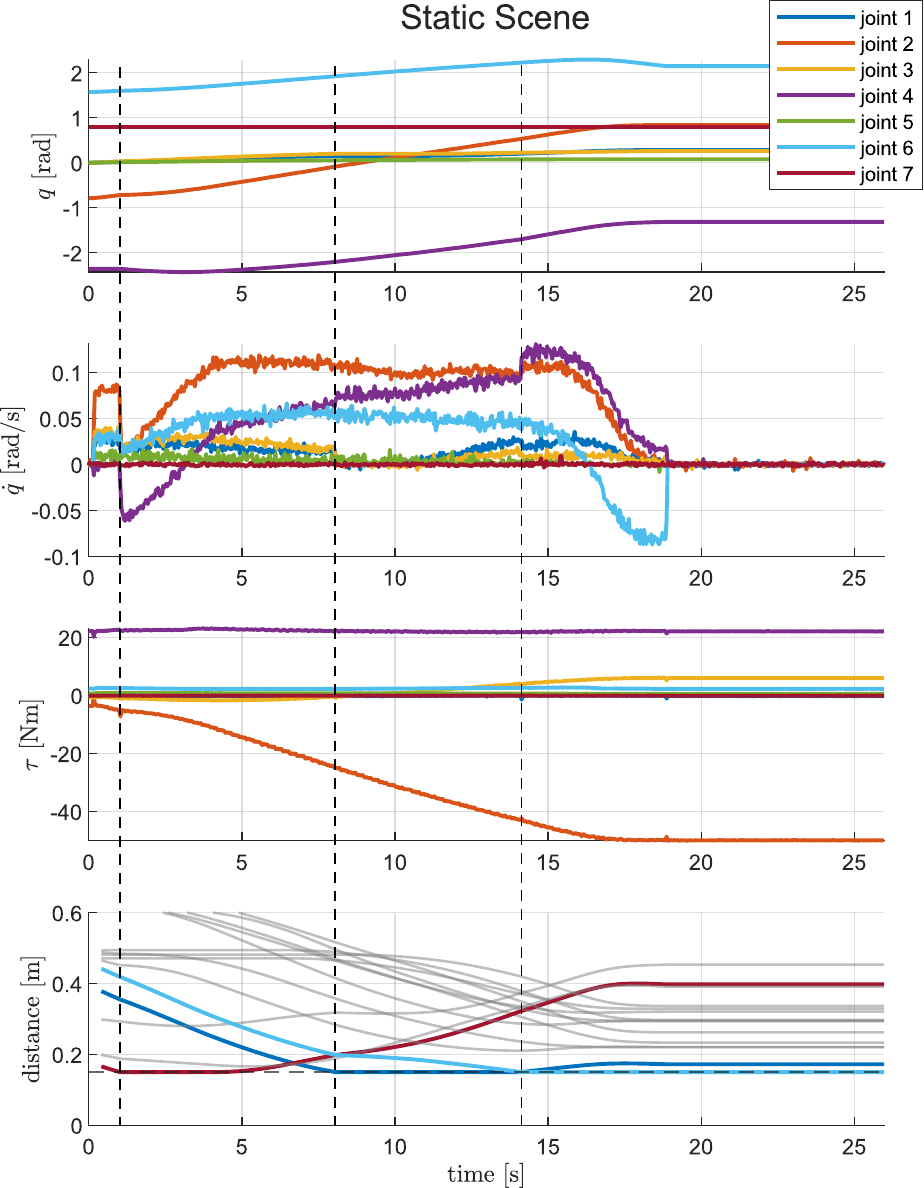}
        \caption{Real robot joint position, velocity, torque, and link-obstacle distances for static scene. Distances not triggering safety boundary (horizontal dashed lines on 0.15m) displayed in gray.}
        \label{fig:static-data}
    \end{subfigure}

    \begin{subfigure}[b]{0.45\textwidth}
        \centering
        \includegraphics[width=\textwidth]{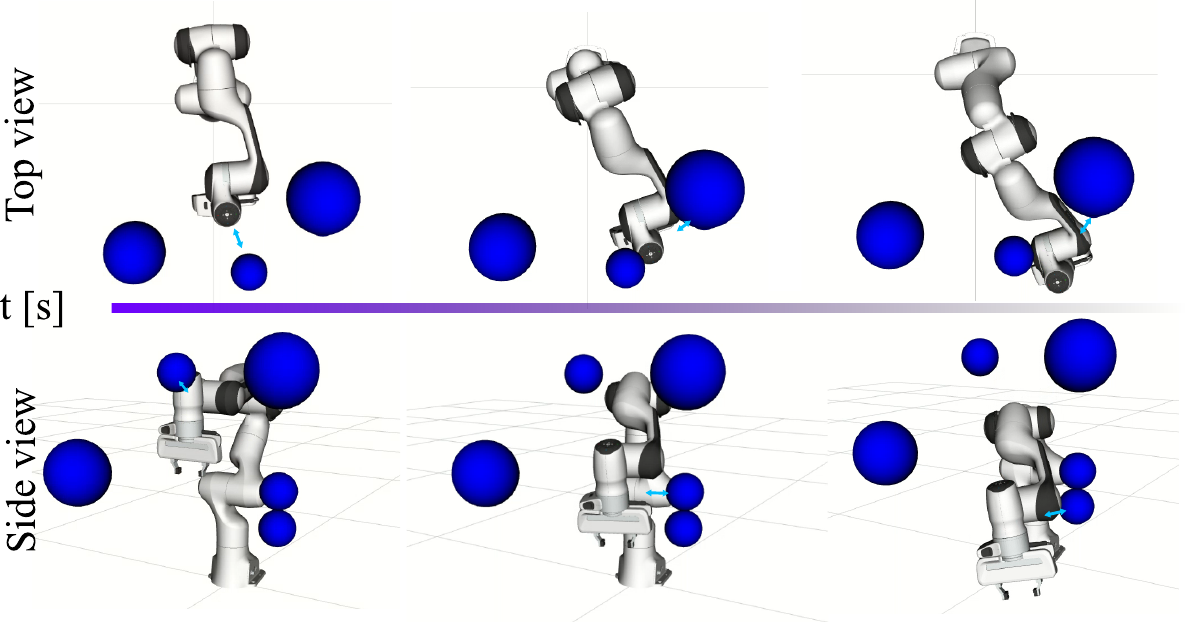}
        \caption{Three critical moments for the static scene (cyan arrow: denotes the particular link and obstacle which triggers minimum safety distance).}
        \label{fig:static-scene}
    \end{subfigure}
    \caption{Data recorded from hardware experiments with static obstacles. (a) the critical moments marked with vertical dashed lines that an obstacle is attempting to violate the safety distance. This triggers the reaction of robot, which can be observed from the robot states as visualized in (b).}
    \vspace{-20pt}
    \label{fig:scene}
\end{figure}

Our first experiment is a scene to reach a goal with $5$ static obstacles. The setup and results can be observed from \cref{fig:static-data,fig:static-scene}. 
As can be seen in \cref{fig:static-data}, we mark the $3$ instances when the complementarity constraints are active (closest obstacle distance reaching the safety threshold of \SI{15}{cm} in the bottom panel) with vertical dashed lines. The expected reactions in the form of unfiltered joint velocities can also be noticed, which stay within the bounds of the system. Similar to the simulation studies, the joint positions also reflect minimal movement unless triggered. Additionally, we plotted the joint efforts (torques) to demonstrate that the continuity constraints are not violated at any point.

\subsection{Scenario \#2: Non-stationary Environment}

\begin{figure}[ht]
    \centering
    \begin{subfigure}[b]{0.45\textwidth}
        \centering
        \includegraphics[width=\textwidth]{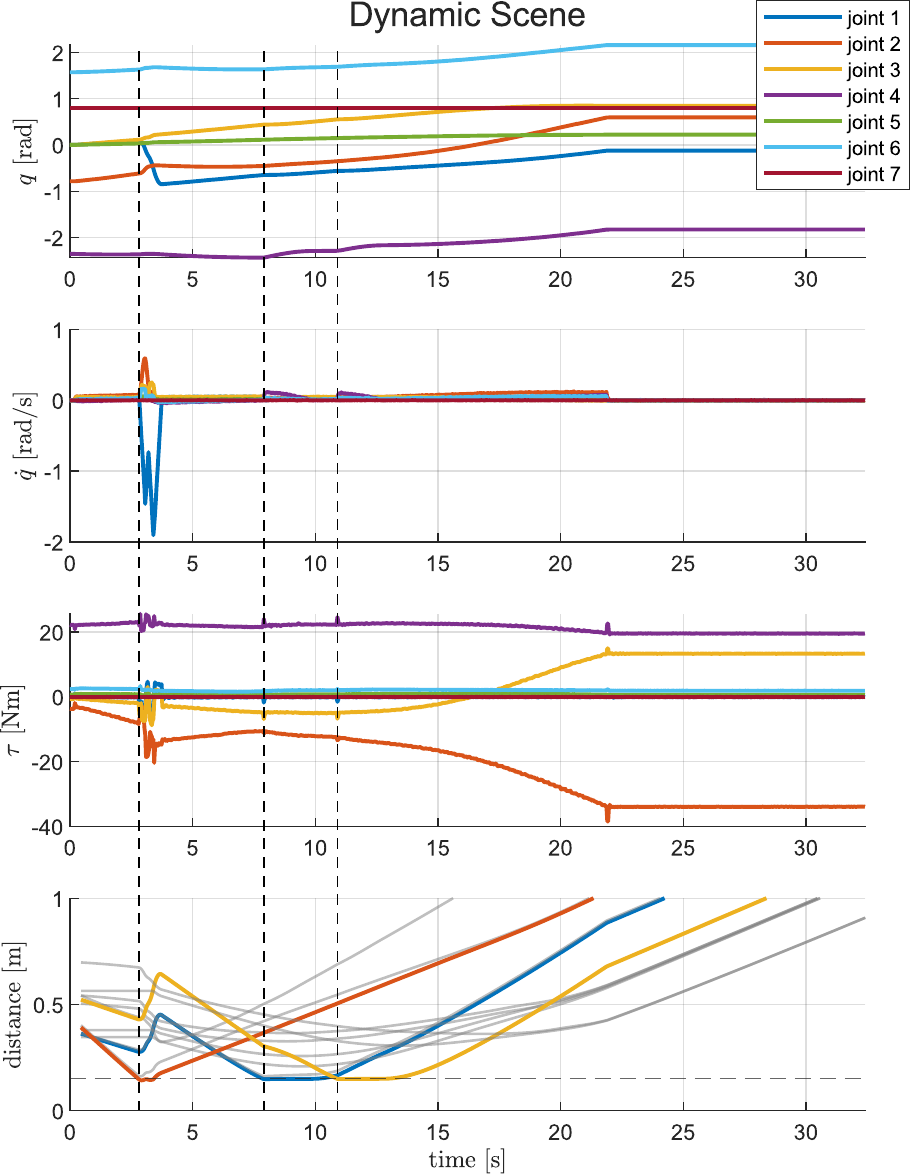}
        \caption{Robot joint position, velocity, torque and link-obstacle distances for dynamic scene. The horizontal dashed lines mean the same as Fig.~\ref{fig:static-data}.}
        \label{fig:dynamic-data}
    \end{subfigure}

    \begin{subfigure}[b]{0.45\textwidth}
        \centering
        \includegraphics[width=\textwidth]{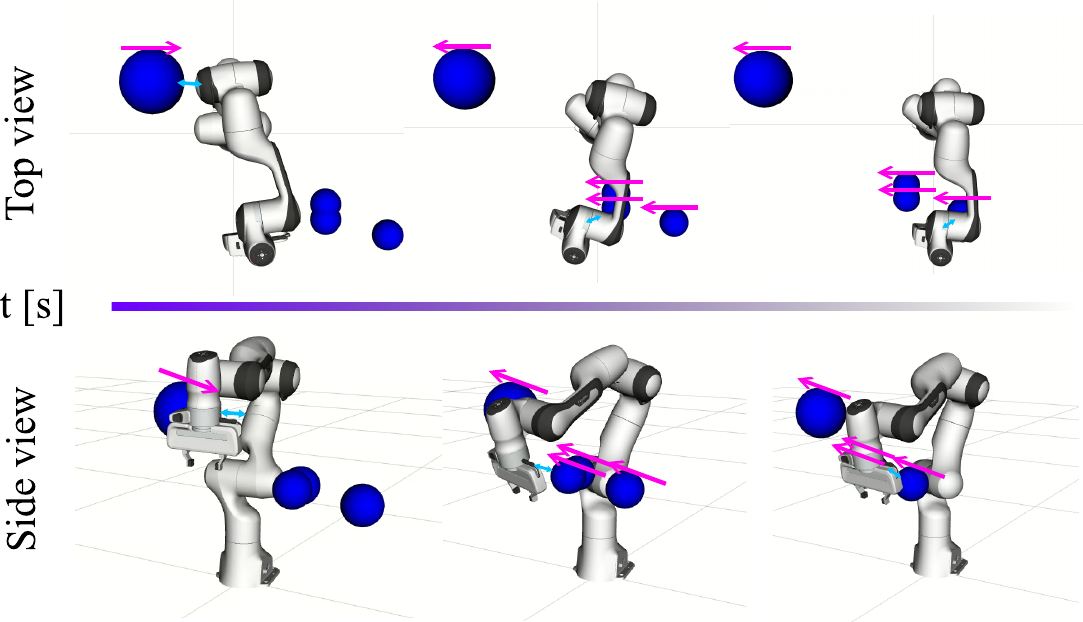}
        \caption{three critical moments for dynamic scene (cyan arrow: same as \cref{fig:static-scene}, magenta arrow: direction of the motion of dynamic obstacles.)}
        \label{fig:dynamic-scene}
    \end{subfigure}
    \caption{Data recorded from hardware experiments with dynamic obstacles. (a) the critical moments marked with vertical dashed lines that an obstacle is attempting to violate the safety distance. This triggers the reaction of robot, which can be observed from the robot states as visualized in (b).}
    \label{fig:scene2}
    \vspace{-30pt}
\end{figure}

The second experiment showcases the ability to handle dynamic, fast-changing scenarios. The analysis in \cref{fig:dynamic-data,fig:dynamic-scene} has a similar structure to the static case. All the obstacles navigate the environment with a velocity of $5$cm/s. As illustrated in the event strip in the bottom panel, the large obstacle first attempts to contact the robot's elbow, which reacts accordingly and then retreats in the opposite direction. This is immediately followed by a group of $3$ obstacles moving from right to left, causing a robot reaction in the hand. Overall, the safety distance is maintained all the time. 

\vspace{-10pt}
\subsection{Scenario \#3: Reaching Task with Interactive Obstacle}

Lastly, we design an ancillary experiment to show the rapid reaction properties of the FLIQC planner. Since integrating a vision pipeline for environment representation and obstacle modeling is not the central focus of this work, we introduce interactive markers as spherical obstacles. 
As shown in Fig.~\ref{fig:inter_marker}, the task for the robot is to reach a target Cartesian pose starting from an initial configuration. We nudge the robot in RViz at different links using the spherical obstacle (controlled by the mouse) during the motion execution. As expected, the real robot swiftly reacts to the stimulus. The arrows in Fig.~\ref{fig:inter_marker} denote the particular link that experiences a change in velocity due to the complementarity constraints. Since the update frequency of the interactive marker is much slower than the controller loop and has a discontinuous jump, we apply a first-order infinite impulse response (IIR) low-pass filter $x[t] = a\cdot x\left[t-1\right] + \left(1-a\right) u$ on the marker positions to smoothen out the obstacle information.

\vspace{-7pt}
\section{Conclusion \& Future Works} \label{sec:conclusion}
\vspace{-3pt}
We introduced a new optimization-based solution for high-frequency motion planning in fast-changing scenarios. Our reactive model, thanks to a robust formalization of linear complementarity constraints, captures surface impenetrability between unstructured environmental obstacles and robot links. Further, it computes feasible joint velocity commands that can directly be used as robot control input without any additional post-processing. We assess the performance of our proposed method with existing planners in an extensive fashion, resulting in noticeable improvement. Real robot experiments further strengthen the applicability of our FLIQC planner in diverse scenes including an interactive scenario. A more robust planner in the ``global" sense can be achieved by incorporating a Model Predictive Control (MPC) based machinery, but inherent uncertainties demand more sophisticated solutions~\cite{mayne2011robust}. Hence, future works will involve integrating a vision pipeline for obstacle tracking with complex geometries and incorporating end-effector constraints for a wide range of applications with global features.      

\bibliographystyle{IEEEtran}
\bibliography{bibliography}

\end{document}